\documentclass[10pt,twocolumn,letterpaper]{article}

\usepackage{wacv}              
\usepackage{times}
\usepackage{epsfig}
\usepackage{graphicx}
\usepackage{amsmath}
\usepackage{amssymb}
\usepackage{booktabs}
\usepackage{url}

\usepackage[utf8]{inputenc}
\usepackage[T1]{fontenc}
\usepackage[normalem]{ulem}
\usepackage{caption}
\usepackage{wrapfig}
\usepackage{algpseudocode} 
\usepackage[ruled,vlined]{algorithm2e}
\usepackage[normalem]{ulem}
\usepackage{multirow, multicol}
\usepackage[pagebackref,breaklinks,colorlinks]{hyperref}

\usepackage[capitalize]{cleveref}
\usepackage{pifont}
\crefname{section}{Sec.}{Secs.}
\Crefname{section}{Section}{Sections}
\Crefname{table}{Table}{Tables}
\crefname{table}{Tab.}{Tabs.}

\def\wacvPaperID{1413} 
\def\confName{WACV}
\def\confYear{2024}

\begin{document}

\title{Partial Binarization of Neural Networks for Budget-Aware Efficient Learning}
\author{
Udbhav Bamba\textsuperscript{\ding{61}}\thanks{Equal Contribution}, Neeraj Anand\textsuperscript{\ding{66}}\footnotemark[1], Saksham Aggarwal\textsuperscript{\ding{61}}\footnotemark[1], Dilip K. Prasad\textsuperscript{\ddag} and Deepak K. Gupta\textsuperscript{\ding{61}}\footnotemark[1] \\
\textsuperscript{\ding{66}} Nyun AI, India \\
\textsuperscript{\ding{61}}Transmute AI Lab, India \\
\textsuperscript{\ddag}UiT The Arctic University of Norway, Norway \\
{\tt\small \{ubamba98,neerajanandfirst,sakshamaggarwal20\}@gmail.com}
}

\maketitle

\begin{abstract}
Binarization is a powerful compression technique for neural networks, significantly reducing FLOPs, but often results in a significant drop in model performance. To address this issue, partial binarization techniques have been developed, but a systematic approach to mixing binary and full-precision parameters in a single network is still lacking. In this paper, we propose a controlled approach to partial binarization, creating a \emph{budgeted binary neural network (B2NN)} with our $\texttt{MixBin}$ strategy. This method optimizes the mixing of binary and full-precision components, allowing for explicit selection of the fraction of the network to remain binary. Our experiments show that B2NNs created using $\texttt{MixBin}$ outperform those from random or iterative searches and state-of-the-art layer selection methods by up to 3\% on the ImageNet-1K dataset. We also show that B2NNs outperform the structured pruning baseline by approximately 23\% at the extreme FLOP budget of 15\%, and perform well in object tracking, with up to a 12.4\% relative improvement over other baselines. Additionally, we demonstrate that B2NNs developed by $\texttt{MixBin}$ can be transferred across datasets, with some cases showing improved performance over directly applying $\texttt{MixBin}$ on the downstream data. \footnote{Code is publicly available at \href{https://github.com/transmuteAI/trailmet}{https://github.com/transmuteAI/trailmet}}
\end{abstract}

\section{Introduction}
Convolutional neural networks (CNNs) have led to several breakthroughs in the field of computer vision and image processing, especially because of their capability to extract extremely complex features from the images. However, these deep CNN models are extremely computation-hungry and require significant power to process. For example, a ResNet-18 classification model comprises 1.9 million parameters, each represented in full-precision using 32-bits, and accounts for a total of 1.8 billion floating point operations (FLOPs) for the ImageNet dataset \cite{ILSVRC15}. For most of the problems, these deep CNN models are overparameterized, and there is enormous scope of reducing their sizes with minimal to almost no drop in the performance of the models. The popular approaches for effective model compression include removing the non-important set of parameters or channels (pruning)  \cite{Liu2017learning}, distilling knowledge of the dense teacher network into a light-weight student network (distillation) \cite{Hinton2015DistillingTK}, converting 32-bit representations of the parameters to lower bit representations (quantization) \cite{Krishnamoorthi2018QuantizingDC}, and transforming the network weights to 1-bit representations (binarization) \cite{Courbariaux2016BinarizedNN}.

\begin{figure}\centering
\centering
\includegraphics[width=\linewidth]{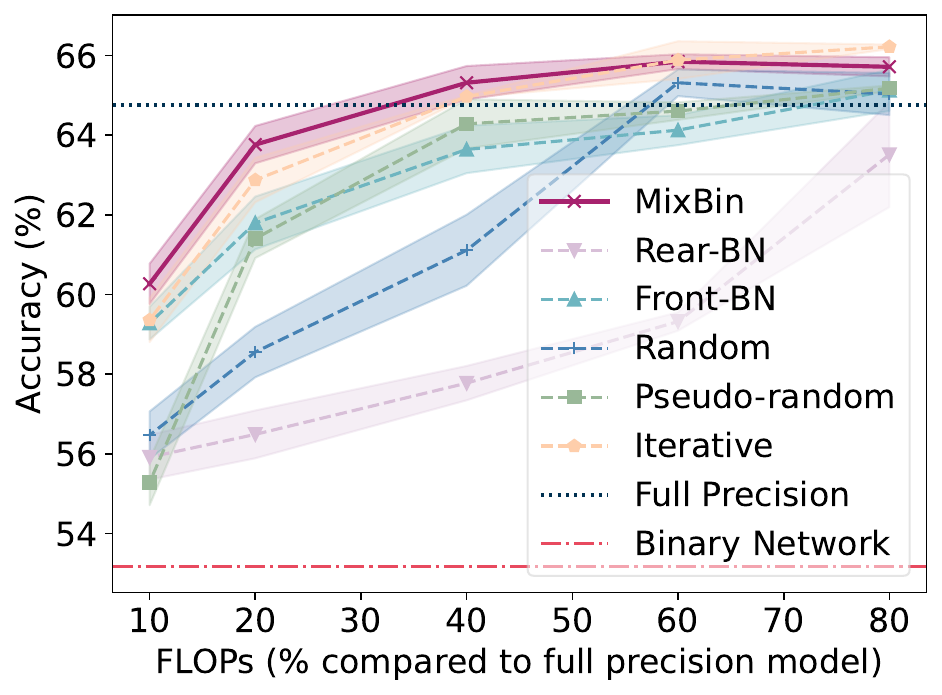}
\caption{\textbf{Performance comparision of B2NNs obtained using \texttt{MixBin} vs. the various trivial approaches}. The B2NNs are constructed using cResNet-20 architecture on CIFAR-100 datasets. Here, \emph{Rear-BN} and \emph{Front-BN} represent B2NNs constructed with binary layer placed in the rear and front parts of the network, respectively. \emph{Pseudo-random} involves modifying 30\% of the MixBin generated B2NNs. The Full Precision network and Binary Network have $4.14\times10^7$ and $1.93\times10^6$ FLOPs respectively.}
\label{fig-binzone}
\end{figure}

Among the methods outlined above, binarization is very effective in drastically reducing the size of the model and increasing the inference speed. In its basic form, binarization involves changing all the weights and activations to 1-bit representation and implementing the convolutions with bitwise XNOR operations \cite{Rastegari2016XNORNetIC}. However, due to the significantly reduced representation, the performance of the binarized network is significantly lower than its full-precision counterpart. To circumvent this, several approaches exist such as binarizing only the weights and keeping the activations as full-precision \cite{Courbariaux2015BinaryConnectTD}, parallely stacking multiple binarized layers \cite{Lin2017TowardsAB}, using special layers such as squeeze-and-excitation blocks \cite{Martnez2020TrainingBN} and retaining skip connections as full-precision modules in a ResNet-type binary network \mbox{\cite{liu2020bi}}. All the above methods exploit partial binarization of the network reduce the performance gap between full precision network and binarization. However, there still does not exist a systematic approach to perform intermediate levels of compression in a more controlled sense. A solution as such would provide the flexibility of analyzing the drop in the performance of the model at different levels of model compression, thereby allowing to choose a right balance between the size of the binararized model and its performance. Beyond this, such an approach would provide the control on using binarization to perform hardware-specific compression, thereby allowing the compressed model to exploit the full computational power budget of the target hardware.

An alternative to using partial binarization could be quantization, where based on the desired extent of compression, the choice of precision for the target network can be made. However, due to its ability to replace the matrix multiplications with XNOR operations, binarization can achieve a higher extent of FLOP reduction compared to an equivalent amount of quantization in terms of similar number of effective parameters. Moreover, binarization itself can be looked at as a modified form of aggressive quantization, and it is worth understanding how it would fair. 


In this paper, we propose a paradigm to perform partial binarization of neural networks in a controlled sense, thereby constructing \textit{budgeted binary neural network (B2NN)}. Our B2NN approach relies on identifying the right set of convolutional layers of a network that should be binarized, and the rest of the network is retained as full-precision. A straightforward approach to select layers to binarize is through random sampling, and we experimentally demonstrate that this is not very effective. While the network itself can be made very light, the performance of the compressed variant deteriorates significantly. Through various trivial baselines, we numerically demonstrate that binarizing different parts of a CNN has very different effect on the performance of the compressed model, thus making it important that the right set of target layers are identified for building the right B2NN. This is shown in Figure \ref{fig-binzone}, and we see that for a similar computational budget, keeping the later layers of a network as full-precision boosts its performance significantly, while more binarization towards of the end of the network architecture has an adverse effect. We discuss this aspect further in section \ref{Experiments} of this paper. 

To overcome the challenge outlined above, we present \texttt{MixBin}, a smart selection strategy that constructs B2NN through optimized mixing of the binary and full-precision components. \texttt{MixBin} allows to explicitly choose the fraction of the network to be kept as binary, thereby presenting the flexibility to approximately adapt the inference cost to a prescribed budget. The core of our selection strategy lies in effectively analyzing how sensitive the performance of the network is with respect to the different layers. Based on our initial observations, we have formulated two variants, \texttt{MixBin}$_{Loss}$ and \texttt{MixBin}$_{Grad}$, and the related details are described in section \ref{sec-mibin} of this paper. We conduct several experiments and demonstrate in several different scenarios that the resultant compressed networks achieved using \texttt{MixBin} are superior over all the baselines chosen in our study, which includes some of the popularly used model compression methods as well.

\textbf{Contributions. }The contributions of this paper can be summarized as follows.

\begin{itemize}
    \item We introduce the concept of budgeted binary neural networks (B2NNs), compressed variants of the dense full-precision models obtained through partial binarization. B2NN relies on effectively identifying the right set of layers that are to be retained as binary or full-precision.
    \item We present \texttt{MixBin}, a search strategy to compress a given full-precision  network through partial binarization in an optimized sense. The inherent design of \texttt{MixBin}  facilitates budgeted binarization, allowing to develop light-weight models that maximally exploit the available compute resources. We demonstrate through numerical experiments that B2NNs obtained from our \texttt{MixBin} strategy are significantly better than those obtained from random selection or even iterative greedy search over the network layers.
    \item The efficacy of \texttt{MixBin} is demonstrated on multiple datasets and tasks for various budget scenarios, and the resultant models obtained from \texttt{MixBin} are consistently superior over those obtained with popular model compression methods chosen as baselines in this study. Performance results on ImageNet demonstrate the effectiveness of \texttt{MixBin} for large datasets, and we also show that it works well for the downstream task of object tracking.

\end{itemize}

\section{Related Work}

Neural networks are often known to be overparameterized \cite{Frankle2019TheLT}. This leads to undesired additional latency when deploying these models in production with little effect on the model accuracy. There exist several works that focus on addressing this issue. First among these is through inducing efficient components in the neural network design, such as using bottleneck blocks \cite{He2016DeepRL}, replacing the $3\times3$ convolutions with $1 \times 1$ \cite{Iandola2016SqueezeNetAA}, using depthwise-separable convolutions \cite{Howard2017MobileNetsEC} and employing neural architecture search \cite{Zoph2017NeuralAS, Pham2018EfficientNA, Tan2019MnasNetPN}. Another approach to design efficient networks involves distilling the information of large networks into smaller networks (knowledge distillation)  \cite{Hinton2015DistillingTK, Urban2017DoDC, Ba2014DoDN, Romero2015FitNetsHF, Tian2020ContrastiveRD}. Further, there exist works that aim at identifying the undesired or less desired weights or filters of a network and removing them. Selection criteria for pruning include ranking scores based on L1/L2 norms of parameters \cite{Li2017PruningFF, He2017ChannelPF, Frankle2019TheLT}, gradients derivatives \cite{Dong2017LearningTP, Lee2019SNIPSN}, learnable parameters \cite{Lemaire2019cvpr, tiwari2021chipnet, Liu2017learning} and pruning of network once at intialization prior to training \cite{lee2018snip, tanaka2020pruning} . An equally effective direction of efficient network design is through quantization of the weights and/or activations of a network to represent it with a reduced number of bits \cite{Krishnamoorthi2018QuantizingDC, Banner2018ACIQAC, Zhao2020LinearSQ, Gholami2022ASO}.

Quantizing a given network refers to representing each weight or activation with reduced number of bits, such as half-precision (16-bits) or 8-bits. An extreme case of quantization is network binarization where the 32-bit representations are directly scaled down to 1-bit each \cite{Courbariaux2016BinarizedNN}. Due to the replacement of the conventional convolution operation with a bitwise XNOR operation, a significant computational gain is observed. Further, adding the channelwise scaling on the binary weights allows to scale BNNs to largescale datasets such as ImageNet \cite{Rastegari2016XNORNetIC} . Although binarization is an effective technique for network compression, it leads to a significant reduction in network representation which results in a significant decrease in performance. There exist several works that attempt to find a right balance between model performance and the extent of compression in the model. For example, ABCNet proposes to stack multiple parallel layers together to use multiple binary layers to increase the representation capability of the network \cite{Lin2017TowardsAB}. In BiRealNet \cite{liu2020bi}, skip connections are represented as real-valued and it has been shown to boost the performance. Other recent binarization methods that improve performance include Reactnet \cite{Liu2020ReActNetTP}, IR-Net \cite{Qin2020ForwardAB} and SA-BNN \cite{Liu2021SABNNSB}.

Most of the approaches listed above attempt to find representations in between a fully binary network and real-valued one. However, there is no straightforward method to design networks comprising binary components that make efficient use of the available computational memory and delivering maximum possible performance. The closest towards this goal is the hybrid binary network \cite{8354199} that performs selective binarization through locally converting the activations to full-precision and retaining the rest of the network as binary . However, this approach provides limited flexibility in terms of full exploitation of the mixing between binary and full-precision components. There also exists layer selection methods for quantization designed to combine different precisions layers together in a network \cite{Chen2019BinarizedNA, 9577892, NEURIPS2020_d77c7035}. These use either the difference of network weights, their gradients or the Hessians to decide which layers to choose for reduced precision and the extent of reduction. We discuss these methods further in Section \ref{Experiments} and present a comparison of these methods with \texttt{MixBin}.










\section{Proposed Approach}

\subsection{Background}
Binary neural networks (BNNs), also referred to as 1-bit neural networks, use binary weight parameters and binary activations for the intermediate layers of the network, excluding the first layer. $\text{Sign}(\cdot)$ function is used to convert real-valued weights/activations to their binary counterparts and the conversions can be mathematically stated as

\begin{align}
    \begin{aligned}
    a_b = \text{Sign}(a_r) = 
        \begin{cases}
        -1 \text{ if }a_r < 0 \\
        +1 \text{ otherwise}
        \end{cases} \\
    w_b = \text{Sign}(w_r) = 
        \begin{cases}
        -1 \text{ if }w_r < 0 \\
        +1 \text{ otherwise}
        \end{cases}
    \end{aligned}
\label{eq-sign}
\end{align}

where $a_r$ and $w_r$ denote the real-valued (full-precision) activations and weights, and $a_b$ and $w_b$ the corresponding binary variants.

Compared to the full-precision model where 32-bit representations are used for every parameter, BNNs, with their 1-bit representations, can lead up to $32\times$ memory saving. Further, since the activations are also chosen as binary, the convolution ($*$) operation is implemented as a bitwise XNOR ($\oplus$) operation and a bit-count operation. It is represented as
\begin{equation}
    \mathbf{a}_r * \mathbf{w}_r \approx \boldsymbol\alpha \odot (\mathbf{a}_b \oplus \mathbf{w}_b)
\end{equation}

where $\boldsymbol\alpha \in \mathbb{R}_+^{c_{out}}$  contains the channelwise scaling factors and $\odot$ denotes the elementwise multiplication operation. For $\mathbf{w}_r \in \mathbb{R}^{c_{\text{out}} \times c_{\text{in}} \times k_{\text{h}} \times k_{\text{w}}}$, the scaling factor $\alpha_i \in \boldsymbol\alpha$ can be mathematically represented as
\begin{equation}
    \alpha_i = \frac{1}{n} \sum \mathbf{w}_r^{(i,:,:,:)}
\end{equation}
denoting summation of the matrix along all dimensions except $c_{\text{out}}$, and $n=c_{in}\times k_h \times k_w$. Here, $c_{in}$, $k_h$ and $k_w$ denote the input channel dimension, kernel height and width, respectively. For more details related to the scaling, see \cite{Rastegari2016XNORNetIC}.

\subsection{Budgeted binary neural network (B2NN)}
Although BNN leads to significant memory and computation gain, it has been experimentally demonstrated that the performance of BNNs can be significantly lower than their full-precision counterparts. Clearly, reducing 32-bits to 1-bit in all parts of the network is not the effective way, and budgeted binary neural network alleviates this issue. Among the various layers of a given CNN for example, B2NN identifies the right set of layers that are to be converted to 1-bit representation with minimal compromise in model performance, and the rest are retained as full-precision. Below we provide the mathematical description of B2NN.

Let $\mathcal{F}: \mathbb{R}^d \rightarrow \mathbb{R}^c$ denote a neural network comprising a set of weights $\mathbf{W}$, activations $\mathbf{A}$, and input and output layers. For $\mathcal{F}$ comprising $n$ hidden layers, this implies, we have \mbox{$\mathbf{W} = \{\mathbf{w}_1, \mathbf{w}_2, \hdots, \mathbf{w}_n, \mathbf{w}_{n+1}\}$} and \mbox{$\mathbf{A} = \{\mathbf{a}_1, \mathbf{a}_2, \hdots, \mathbf{a}_n\}$}. Note here that $\mathbf{w}_{n+1}$ performs a fully-connected mapping between the output of the final convolutional layer and the output of the model. During binarization, it is common for even fully-binarized networks to retain $\mathbf{w}_1$, $\mathbf{w}_{n+1}$ and $\mathbf{a}_{n}$ as full-precision \cite{Liu2020ReActNetTP, liu2020bi}, and we follow a similar convention . Thus, for methods that perform complete binarization, $\mathbf{w}_i \enskip \forall \enskip i \in [2, n]$ and $\mathbf{a}_i \enskip \forall \enskip i \in [1, n-1]$ are converted from full-precision to binary. However, as stated earlier, this dips the performance of the resultant BNN significantly, and alternatively some of the works retain $\mathbf{a}_i$ as full-precision.

B2NN couples $\mathbf{a}$ and $\mathbf{w}$ together as $\boldsymbol\theta_i = \{\mathbf{a}_i, \mathbf{w}_{i+1}\}$, referring them as one layer, and performs binarization on a subset \mbox{$\boldsymbol\Phi \subset \boldsymbol\Theta$}, where \mbox{$\boldsymbol\Theta = \{\boldsymbol\theta_1, \boldsymbol\theta_2, \hdots, \boldsymbol\theta_{n-1}\}$}. The generic mathematical problem that we solve with B2NN can be stated as follows.

\begin{equation}
    \begin{aligned}
        \boldsymbol\Phi^* = \underset{\boldsymbol\Phi \subset \boldsymbol\Theta, \mathbf{W}}{\text{argmin}} \enskip \mathcal{L}(\mathcal{F}(\boldsymbol\Phi, \boldsymbol\Theta-\boldsymbol\Phi, \mathbf{x}), \mathbf{y}) \quad \\
        \text{s.t.} \quad \mathcal{B}(\boldsymbol\Phi, \boldsymbol\Theta-\boldsymbol\Phi)\leq \mathcal{B}_0,
    \end{aligned}
\label{eq-b2nn}
\end{equation}

where $\boldsymbol\Phi^*$ denotes the optimized subset of layers that are binarized, and $\mathcal{L}(\cdot)$ denotes the function to be minimized on the dataset $(\mathbf{x}, \mathbf{y})$ when making this selection. Further, $\mathcal{B}(\cdot)$ denotes the budget function and $\mathcal{B}_0$ is the prescribed limit. In this paper, we use FLOPs budget for compressing the networks.

\textbf{Effect of binarizing different parts of a network. }
\label{sec:effect_of_binarization}
The performance of the constructed B2NN model depends heavily on the choice of $\boldsymbol\Phi$. To demonstrate the importance of selection, we consider several trivial approaches in this study to construct B2NNs, where the goal is to select $k$ out of $n-1$ layers to be binarized, such that the performance of the resultant B2NN is maximized to the largest possible extent while also satisying the budget constraint as stated in Eq. \ref{eq-b2nn}. We provide a brief description of the various trivial approaches below followed by detailed formation of our \texttt{MixBin} approach in Section \ref{sec-mibin}.

\emph{Random selection. }As the name suggests, this approach does not involve any smart strategy in the selection and the process of selecting $k$ out of $n-1$ layers to be binarized is completely random.

\emph{Front- or Rear-BN selection. }In this selection strategy, $\boldsymbol\Phi^*$ is sampled sequentially from the front or rear parts of the full-precision network, respectively. For Front-BN selection, this implies selecting $\boldsymbol\Phi^*$ as \mbox{$\boldsymbol\Theta = \{\boldsymbol\theta_1, \boldsymbol\theta_2, \hdots, \boldsymbol\theta_{k}\}$}. Similarly, for Rear-BN selection , we have \mbox{$\boldsymbol\Theta = \{\boldsymbol\theta_{n-k}, \boldsymbol\theta_{n-k+1}, \hdots, \boldsymbol\theta_{n-2}, \boldsymbol\theta_{n-1}\}$}.

\emph{Iterative selection. }It is a greedy selection process based on iterative search strategy designed to identify the right layers of any given network to be converted to binary or full precision, one-by-one. For the $k$ out of $n-1$ layers to be binarized, the $j^{\text{th}}$ step of binarization, where $j \in [1, k]$, can be stated as finding the optimal layer $\boldsymbol\theta^* \in \boldsymbol\Theta_{(j)}$ to be binarized. It can be mathematically stated as:
\begin{equation}
\begin{split}
    \boldsymbol\theta^*_{(j)} = \underset{\boldsymbol\theta \subset \boldsymbol\Theta_{(j)}, \mathbf{W}}{\text{argmin}} \enskip \mathcal{L}(\mathcal{F}(\boldsymbol\theta+ \boldsymbol\Phi_{(j)}, \boldsymbol\Theta_{(j)}-\boldsymbol\theta, \mathbf{x}), \mathbf{y}) \enskip \\ 
    \text{s.t.} \quad \mathcal{B}(\boldsymbol\theta+ \boldsymbol\Phi_{(j)}, \boldsymbol\Theta_{(j)}-\boldsymbol\theta)\leq \mathcal{B}_0,
\end{split}
\end{equation}

where $\boldsymbol\Theta_{(j)} = \boldsymbol\Theta - \boldsymbol\Phi_{(j)}$. 
Here, $\boldsymbol\Phi_{(j)}$ denotes the layers that have already been binarized in the previous $j-1$ steps and is defined as $\boldsymbol\Phi_{(j)} = \{\boldsymbol\theta^*_{(1)}, \boldsymbol\theta^*_{(2)}, \hdots, \boldsymbol\theta^*_{(j-1)}\}$, where $\boldsymbol\theta^*_{j}$ denotes the optimal layer chosen at the $j^{\text{th}}$ step of binarization to obtain B2NN. For calculating $\boldsymbol\theta^*_{j}$, we perform brute search over all elements of $\boldsymbol\Theta_{(j)}$ and choose the layer, which when binarized, maximizes the performance of the intermediate B2NN model. Additional details and the pseudo-code related this approach can be found in the supplementary material.

\emph{Pseudo-random selection. }This selection strategy involves identifying the $\boldsymbol\Phi_{(j)}^*$ layers to be binarized and then rather than binarizing these layers, a subset of them is replaced by another subset of same size randomly sampled from $\boldsymbol\Theta-\boldsymbol\Phi$. It can be expected that if the process of selecting $\boldsymbol\Phi_{(j)}^*$ is properly optimized with respect to performance of the resultant B2NN, the performance of the network obtained from pseudo-random selection would be relatively sub-optimal.

As shown in \mbox{Figure \ref{fig-binzone}}, mixing full-precision and binary layers using different strategies yields very different results, and a wrong selection approach can lead to completely sub-optimal model performance. For example, the performance of Rear-BN is far inferior to even the random selection strategy, indicating that the later parts of the network should not be preferred for binarization. Among the various methods described above, we observe that Iterative Selection works the best, although being lower than our \texttt{MixBin} approach. However, the iterative selection approach is computationally very expensive which limits its adoption for large networks. For an extremely low budget of 10\%, we observe that almost all the methods result in poorly performing networks, and Pseudo-random ranks the lowest. Clearly, the trivial layer selection strategies described above are not well suited for constructing B2NNs, especially when it comes to extreme model compression. This shows the importance and need of a systematic way to construct B2NNs such that the maximal performance of the original network can be retained. 


\begin{algorithm}
\SetAlgoLined
\SetKwInOut{Output}{Output}
\SetKwInOut{Given}{Given}
\Given{Budget value $\mathcal{B}_0$;\\
        Pre-trained network weight $\boldsymbol\Theta$; \\
        Neural Network $\mathcal{F}$; Training data $\mathcal{D}$;\\
       }
\Output{Binarized weight $\boldsymbol\Phi$} 
$K \gets \{\}$; \\
\For{$\boldsymbol\theta \in \boldsymbol\Theta$}{
    $(\mathbf{x}, \mathbf{y}) \gets \mathcal{D}$; \\
    \tcp{Binarize both activation and weights of $\boldsymbol\theta$}
    $\hat{\boldsymbol\theta} \gets \textbf{\texttt{Binarize}}(\boldsymbol\theta)$; \\
    \tcp{Pass $\mathbf{x}$ through $\mathcal{F}$, B2NN with a single binary layer $\tilde{\boldsymbol\theta}$}
    $\tilde{\mathbf{y}} \gets \mathcal{F}(\hat{\boldsymbol\theta}, \boldsymbol\Theta - \boldsymbol\theta, \mathbf{x})$ \\
    \tcp{Calculate Binarization coefficient for $\boldsymbol\theta$}
    $\kappa \gets \textbf{\texttt{MixBin}}(\mathbf{y}, \tilde{\mathbf{y}}, \hat{\boldsymbol\theta}, \boldsymbol\Theta - \boldsymbol\theta)$ \\
    \tcp{Add tuple ($\kappa, \boldsymbol\theta$) to set $K$}
    $K \gets \textbf{\texttt{push}}(\{\kappa, \boldsymbol\theta\})$  \\
}
\tcp{Sort set $K$ based on $\kappa$}
$K \gets \textbf{\texttt{Sort}}(K | \kappa)$  \\
\tcp{Select $\boldsymbol\Phi$ from $K$ for the prescribed budget $\mathcal{B}_0$}
$\boldsymbol\Phi \gets \textbf{\texttt{Select}}(K | \mathcal{B}_0)$ \\

\caption{\texttt{MixBin} Approach }
\label{mixbin_algo}

\end{algorithm}

\subsection{MixBin}
\label{sec-mibin}
MixBin refers to the strategy of mixing binary and full-precision components in a network in an efficient and effective manner. The approach of MixBin is described in Algorithm \ref{mixbin_algo}. Unlike the greedy selection approach discussed earlier, \texttt{MixBin} identifies the set of layers to be binarized, $\boldsymbol\Phi^*$, in one single pass eliminating the requirement for any model updates via backpropagation. It involves first calculating the binarization coefficient $\kappa$ with respect to every layer. To compute $\kappa_i \enskip \forall \enskip \kappa_i \in [1,  n-1]$, only the $i^{\text{th}}$ layer is converted to binary and the performance of the resultant network is computed. Note that $\kappa$ can be any generic function such that magnitude of $\kappa_i$ correlates with the extent to which the $i^{\text{th}}$ layer would be a preferable choice for binarization. In this paper, we choose it to be a function of drop in performance of the network due to binarization of the $i^{\text{th}}$ layer and/or the change in the gradient of the loss with respect to this layer. 

For the selection of $\boldsymbol\Phi^*$ as described in Eq. \ref{eq-b2nn}, the values of $\kappa$ associated with all the layers are analyzed and a selection is made. In this paper, we present two variations of $\texttt{MixBin}$ that differ how $\kappa$ is calculated. These are described below.




\textbf{MixBin$_{Loss}$} The premise of this formulation assumes that if a specific layer is favored over the others for binarization, converting this layer from full-precision to binary, while maintaining all other layers at full-precision, would result in minimal performance degradation. In other words, the drop in performance of the model due to binarization of a certain layer can be considered independent of the others and can be used as a direct measure of how preferred a certain layer is for binarization. Based on this, the binarization coefficient is defined as
\begin{align}
    \kappa^{Loss}_j &= \mathcal{L}(\mathcal{F}(\boldsymbol\Theta, \mathbf{x}), \mathbf{y})\nonumber - \\
    & \mathcal{L}(\mathcal{F}({\hat{\boldsymbol\theta}_j}, \boldsymbol\Theta-\hat{\boldsymbol\theta}_j, \mathbf{x}), \mathbf{y})
\end{align}

The resultant values of $\kappa$ are then sorted in ascending order and the layers corresponding to the first $k$ sorted values of $\kappa$ are chosen ensuring that the extent of binarization complies with the predefined budget defined for the B2NN.


\textbf{MixBin$_{Grad}$} This formulation is based on the hypothesis that beyond the drop in performance of the model, the inertness of the resultant network also plays an important role in deciding the importance of the layers for binarization. Based on this, we also include the $L_1$ norm of the gradient of the resultant loss with respect to the parameters of the $j_{th}$ layer. The resultant binarization coefficient $\kappa^{Grad}$ can be stated as
\begin{align}
    \boldsymbol\kappa^{Grad}_j =  \boldsymbol\kappa^{Loss}_j \times \left\lVert{\mathcal{L}(\mathcal{F}({\hat{\boldsymbol\theta}_j}, \boldsymbol\Theta-\hat{\boldsymbol\theta}_j, \mathbf{x}), \mathbf{y})}  \cdot  \hat{\boldsymbol\theta_j}\right\lVert
\end{align}
Intuition for the addition of the gradient term is that once a layer has been binarized, it should have have lower tendency to influence the performance of the model, thereby exhibit increased inertness.


\section{Experiments}\label{Experiments}
\subsection{Experimental Setup}
We demonstrate the efficacy of \texttt{MixBin} on the tasks of image classification and object tracking. For image classification, we conduct experiments to compress CIFAR-ResNet20 (referred further as cResNet-20) on CIFAR-100, MobileNet and ResNet-18 on TinyImageNet and ResNet-18 on ImageNet-1K dataset. For Object tracking, we use ResNet-18 model, and train and compress the tracker on GOT-10K dataset. The choice of smaller architectures like cResNet-20 and MobileNet help us to ensure that the architectures do not overfit on simpler datasets like CIFAR-100 and TinyImageNet. Further the choice of ImageNet-1K and GOT-10K help to demonstrate that \texttt{MixBin} is scalable to complex datasets.

\textbf{FLOPs calculation. }We calculate FLoating point OPerations (FLOPs) based on the code publicly available at {\url{https://github.com/Swall0w/torchstat/}}. For the calculations, batch size of 1 is assumed. For B2NNs, total FLOPs of the network is equal to the sum of FLOPs of the full-precision layers and BOPs (Binary OPerations) for binary counterpart. BOPs are determined using the methodology introduced in BiRealNet \cite{liu2020bi}. This involves dividing FLOPs by 64 since modern CPUs can execute bitwise XNOR operations and bit-counting concurrently in groups of 64.

\subsection{Baselines}
\label{sec-baselines}
To demonstrate the efficacy of \texttt{MixBin}, we set up multiple competitive baselines. These include the random and greedy selection strategies as described in Section \ref{sec:effect_of_binarization}, as well as adaptations of works which propose methods for optimal selection of layers based on different bit-widths. We adapt these methods for our use case as baselines in which we either keep the whole layer binary or full precision. Note that while we explain the methods below in terms of quantization, as in their original formulation, we use the resultant indicators to identify the right layers for binarization.

\emph{BNAS } \cite{Chen2019BinarizedNA}. This approach incorporates amplitude loss function as part of the optimization process. This loss function is formulated as the $L_2$ difference between the full precision weights and the binary weights. We employ the amplitude loss as a criterion for selecting layers in the network, where a lower value indicates a higher inclination towards binarization.

\emph{Adaptive MBQ } \cite{9577892}. This approach uses Taylor expansion to build a metric for quantifying the loss sensitivity introduced by quantization. The metric guides on the extent to which each layer should be quantized, and a lower value of their proposed metric indicates a higher level of quantization. 

\emph{HAWQ-V2 }\cite{NEURIPS2020_d77c7035} uses the sensitivity of Hessian trace of weights, scaled by a perturbation of quantization. By considering second-order information stored in the Hessian, it identifies layers that are more sensitive to quantization. A lower value of the metric indicates a higher level of quantization. 

\emph{Network Slimming }\cite{Liu2017learning}. 
Additionally, we also compare our method with a popular network pruning approach, called network slimming \cite{Liu2017LearningEC}. This approach involves adding an $L_1$ penalty term on the scaling parameters of each Batch Normalization layer which induces sparsity in the network.



\begin{figure}
     \centering
     \begin{subfigure}[b]{0.45\textwidth}
         \centering
         \includegraphics[width=\textwidth]{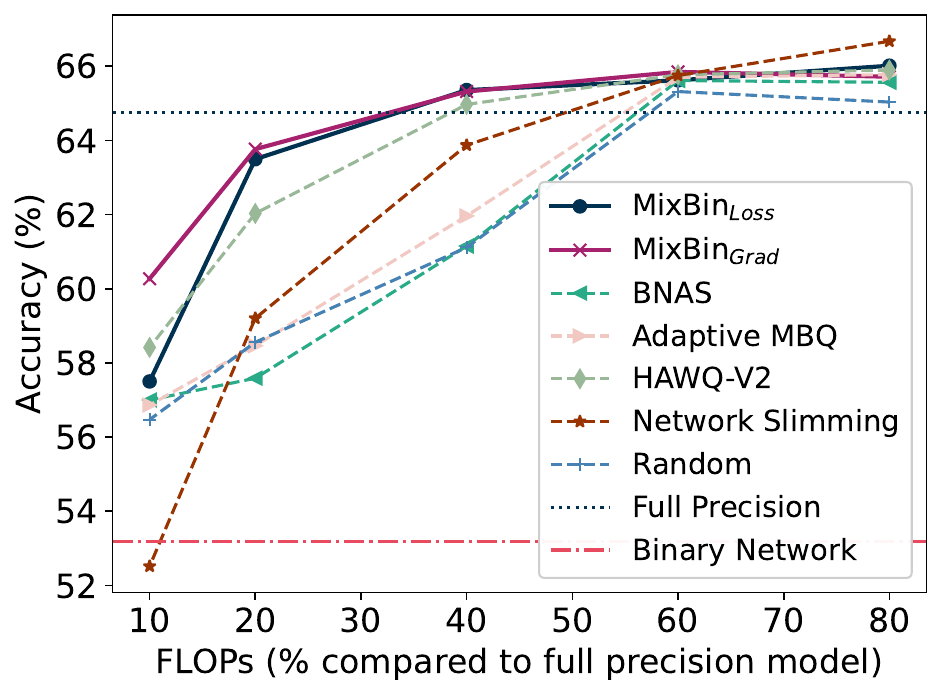}
         \caption{cResNet-20 / CIFAR-100}
     \end{subfigure}
     \hfill
     \begin{subfigure}[b]{0.45\textwidth}
         \centering
         \includegraphics[width=\textwidth]{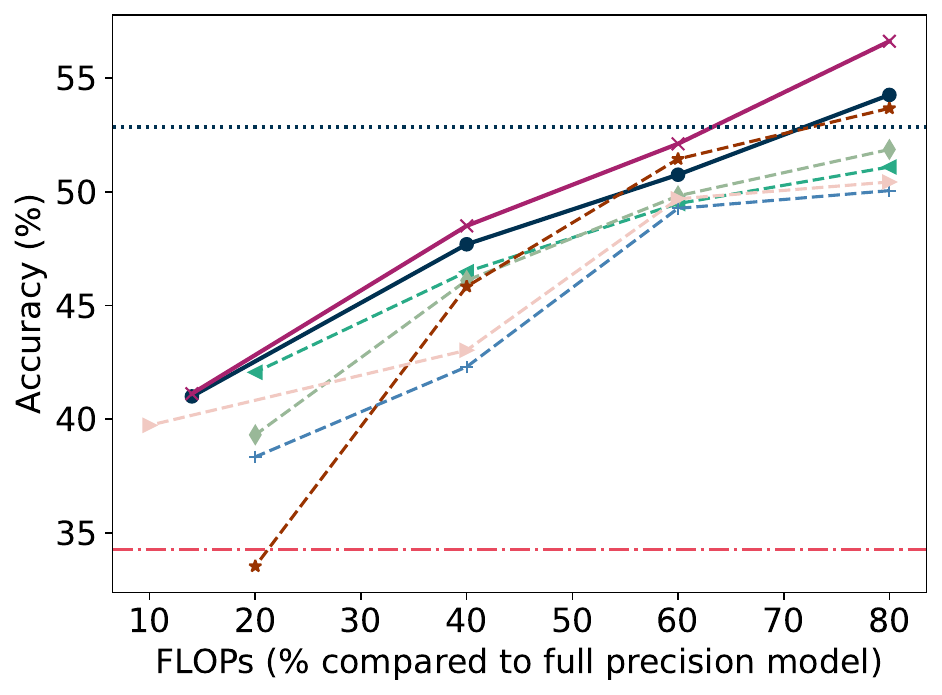}
         \caption{MobileNet / TinyIN}
     \end{subfigure}
     \hfill
     \begin{subfigure}[b]{0.45\textwidth}
         \centering
         \includegraphics[width=\textwidth]{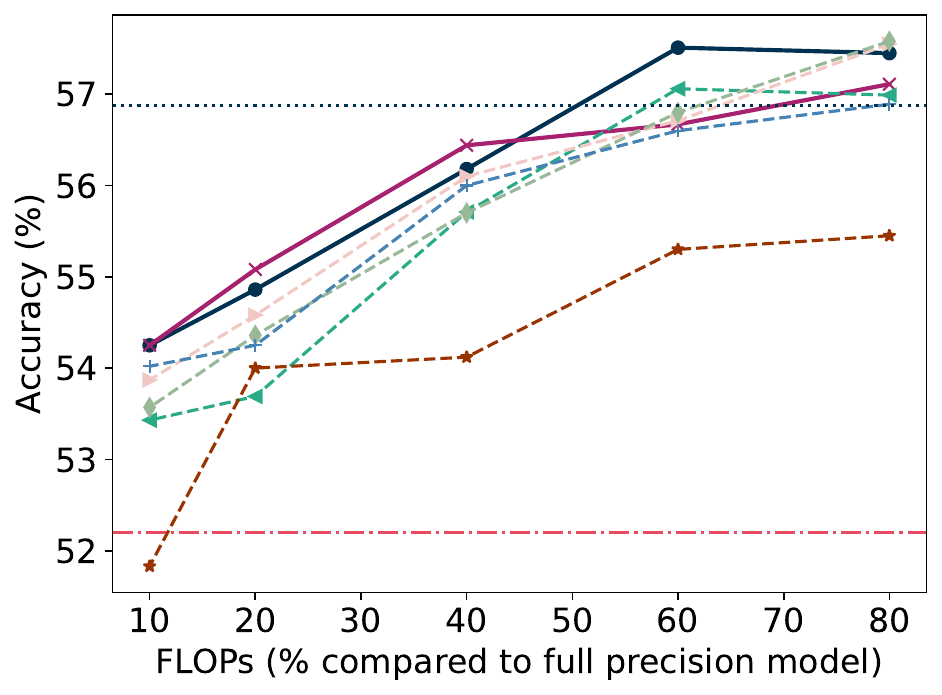}
         \caption{ResNet-18 / TinyIN}
     \end{subfigure}
        \caption{Performance comparison of $\texttt{MixBin}$ with B2NNs obtained using random and competitive layer selection algorithms, as well as pruning, for three different model and data combinations.}
        \label{fig:mixbin_classification}

\end{figure}

\setlength{\tabcolsep}{2pt}
\begin{table}
\caption{Performance scores for  the ResNet-18 architecture on ImageNet-1K datasets for three compression methods: Network Slimming, HAWQ-V2, and \texttt{MixBin} (ours). The Full precision network has a total of $1.82\times10^9$ FLOPs, while the binary network generated using the Bi-RealNet method has $1.71\times10^8$ FLOPs. The term "budget" refers to the percentage of FLOPs that remain after applying the respective compression method.}

\label{table:imagenet_results}
\centering
\resizebox{\columnwidth}{!}{
\begin{tabular}{lcccc}

\toprule
Method & Budget(\%)  &  Acc.(\%)$\uparrow$  & FLOPs(\%)$\downarrow$ &  \\ 
\midrule
Full Precision Network    & - & 65.66 & 100 \\
Binary Network            & - & 55.54 & 9.39 \\
\midrule
Network Slimming          &\multirow{4}{*}{50} & 63.58 & 50\\
HAWQ-V2                   & & 62.25 & 53.16\\
$\texttt{MixBin}_{Grad}$  & & 64.12 & 53.16\\
$\texttt{MixBin}_{Loss}$  & & \textbf{65.38} & \textbf{53.16}\\

\midrule
Network Slimming          &\multirow{4}{*}{15} & 34.58 & 15\\
HAWQ-V2                   & & 56.20 & 15.64\\
$\texttt{MixBin}_{Grad}$  & & 57.26 & 15.64\\
$\texttt{MixBin}_{Loss}$  & & \textbf{57.44} & \textbf{15.64}\\
\bottomrule
\end{tabular}
}

\end{table}



\subsection{Results}
\textbf{Performance on simple classification datasets. }Figure \ref{fig:mixbin_classification} presents the performance curves obtained with \texttt{MixBin} as well as the chosen model compression baselines on CIFAR-100 and TinyImageNet datasets. Baselines here include those mentioned in Section \ref{sec-baselines}, as well as random selection. Across all combinations of the datasets and models, \texttt{MixBin} consistently outperforms all the baselines. At higher budgets, performance achieved with pruning are comparable to those achieved by \texttt{MixBin}. However, when low budgets are used, the performance of the former deteriorates to a level even lower than that of the binary model. Interestingly, we observed that \texttt{MixBin} at budgets of 60\% or higher yields results that are comparable to, or even superior to, the original full precision model. This clearly shows that \texttt{MixBin} leads to compressed models that generalize better, leading to reduced overfitting and thereby improved performance on the evaluation set.

\textbf{Performance on ImageNet-1K. }To further demonstrate the efficacy of \texttt{MixBin}, we study its compression capability on large-scale classification dataset of ImageNet-1K using ResNet-18 architecture. We compare our results to those obtained by HAWQ-V2 and Network Slimming at budget levels of 15\% and 50\%. The performance of each method, along with the corresponding achieved FLOPs ratio compared to the full precision network, are presented in Table \ref{table:imagenet_results}. Pruning exhibited a significant decline in scores at lower budgets, whereas binarization did not experience such a drop. Among the evaluated methods, $\texttt{MixBin}_{Loss}$ demonstrated the highest performance, followed by $\texttt{MixBin}_{Grad}$ and HAWQ-V2. Interestingly, $\texttt{MixBin}_{Loss}$ was able to maintain performance comparable to the full precision network at 50\% budget, while the performance of the other methods dropped by 2-3\%. 

Note that FLOPs of B2NNs are not always met with respect to the corresponding budget since they operate at layer level, unlike structured pruning methods. However, this characteristic can be advantageous for B2NNs as they have a more structured nature and are easy to implement compared to the structured pruning methods. Further, FLOPs of two methods can be exactly the same, while their performance scores may differ due to the presence of multiple layers with similar FLOPs. However, if any one of these binary layers is interchanged from the set of layers having equal FLOPs, it may negatively impact B2NNs' performance.



 

\textbf{Building light weight object trackers. }Object tracking is an application domain that benefits among the most from model compression. When deployed on low-power devices, object trackers need to be light-weight and deliver desired inference speed based on the target hardware. In this regard, we demonstrate the application of \texttt{MixBin} to design light-weight object trackers. We analyze the stability of the compressed variants of the ResNet-18 model at various FLOPs budgets and analyze how well it fairs against model compression achieved using other baselines.
\setlength{\tabcolsep}{2pt}
\begin{table}
\caption{Performance scores for SiamFC with ResNet-18 backbone on GOT-10K datasets for four compression methods: Network Slimming, Adaptive MBQ, BNAS and \texttt{MixBin} (ours). The Full precision network has a total of $4.28\times10^8$ FLOPs, while the binary network generated using the Bi-RealNet method $2.00\times10^7$ FLOPs. The term "budget" refers to the percentage of FLOPs that remain after applying the respective compression method.}
\label{table:tracking_results}
\centering
\resizebox{\columnwidth}{!}{

\begin{tabular}{lcccc}
\toprule
Method & Budget(\%)  &  AO$\uparrow$ & $SR_{0.5}$$\uparrow$  & FLOPs(\%)$\downarrow$ \\ 
\midrule
Full Precision Network    & - & 0.251 & 0.242 & 100 \\
Binary Network            & - & 0.189 & 0.173 & 4.68 \\

\midrule
Network Slimming          &\multirow{5}{*}{60} & 0.235 & 0.220  & 60\\
Adaptive MBQ              & & 0.215 & 0.198  & 56.67\\
BNAS                      & & 0.211 & 0.186  & 61.01\\
$\texttt{MixBin}_{Loss}$  & & \textbf{0.237}&  \textbf{0.219} & \textbf{56.57} \\
$\texttt{MixBin}_{Grad}$  & & \textbf{0.240}  & \textbf{0.234}& \textbf{61.01}\\

\midrule
Network Slimming          &\multirow{5}{*}{40} & 0.198 & 0.179  & 40\\
Adaptive MBQ              & & 0.210 & 0.199  & 39.34\\
BNAS                      & & 0.211 & 0.199  & 39.34\\
$\texttt{MixBin}_{Loss}$  & & \textbf{0.235} &  \textbf{0.226} & \textbf{43.67} \\
$\texttt{MixBin}_{Grad}$  & & \textbf{0.236}  & \textbf{0.228} & \textbf{43.67} \\

\midrule
Network Slimming          &\multirow{5}{*}{10} & 0.201 & 0.182  & 10\\
Adaptive MBQ              & & 0.218 & 0.202  & 13.35\\
BNAS                      & & 0.218 & 0.202  & 13.35\\
$\texttt{MixBin}_{Loss}$  & & \textbf{0.226} & \textbf{0.216} & \textbf{13.35} \\
$\texttt{MixBin}_{Grad}$  & & \textbf{0.226} & \textbf{0.216} & \textbf{13.35} \\
\bottomrule
\end{tabular}
}

\end{table}





Table \ref{table:tracking_results} shows the results for various compressed variants of ResNet-18.
From Table \ref{table:tracking_results}, we see that the compressed models obtained from \texttt{MixBin$_{Grad}$} as well \texttt{MixBin$_{Loss}$} are superior than any other choice of compressions. While this difference is small at FLOPs budget of 60\% of the original network, it grows to a large margin at extreme levels of preserving only 10\% FLOPs from the original network, further confirming the applicability and effectiveness of \texttt{MixBin}. Note that in some cases two methods gave exactly the same set of layers which led to similar scores.

\textbf{Transferability of B2NNs models. }The generaliziblity of the B2NNs obtained from \texttt{MixBin} can be assessed based on the extent to which they are transferable across datasets. This implies analyzing how well a model compressed on one dataset performs on another dataset. In this regard, we present results for the scenario of model transfer from ImageNet-1K to TinyImageNet and CIFAR-100 dataset for classification and from ImageNet-1K to GOT-10K for object tracking. These are reported in Table \ref{table:transfer_results}. From the results, we see that the performance of the transferred models is in a similar range as when the model is trained on the base dataset. Clearly, B2NNs designed for large datasets such as ImageNet seem to work well for the base datasets, thus these do not require any additional layer selections. An interesting observation made on the classification tasks is that at budget of as low as 20\% of the original FLOPs, the transferred B2NNs seems to perform slightly better than those trained on the base dataset. Overall, B2NNs obtained from \texttt{MixBin} seem to generalize well across datasets.

\begin{table}
\caption{Performance scores of B2NN masks transferred (Trans.) from the ImageNet-1K dataset to both a classification task on the TinyImageNet [TinyIN] and CIFAR-100 [C100] dataset and an object tracking task on the GOT-10K dataset, using the ResNet-18 architecture. The term "budget" refers to the percentage of FLOPs that remain after applying the respective compression method. The performance metrics evaluated for the classification task are accuracy (Acc.), while the metrics evaluated for the tracking task are average overlap (AO) and success rate at 0.5 overlap ($SR_{0.5}$)}
\centering
\label{table:transfer_results}
\begin{tabular}{lc|cc|cc}
\toprule
& & \multicolumn{2}{c|}{Classification} & \multicolumn{2}{c}{Object Tracking} \\
Method                       & Budget (\%)  &  TinyIN  $\uparrow$  & C100  $\uparrow$ &  AO $\uparrow$ & $SR_{0.5}$ $\uparrow$ \\ 
\midrule
Trans.  & \multirow{2}{*}{60} & 56.73          & 56.27 & 0.232          & 0.217          \\
Base         &                     & \textbf{57.11} & \textbf{56.37} & \textbf{0.240} & \textbf{0.234}\\
\midrule
Trans. & \multirow{2}{*}{40} & 55.67          & \textbf{55.91} & \textbf{0.237} & \textbf{0.232} \\
Base        &                     & \textbf{56.44} & 55.48 & 0.236          & 0.228          \\
\midrule
Trans. & \multirow{2}{*}{20} & \textbf{55.51} & \textbf{54.69} & \textbf{0.220} & \textbf{0.206} \\
Base        &                     & 55.08          & 52.93 & 0.214          & 0.203          \\
\midrule
Trans. & \multirow{2}{*}{10} & \textbf{54.46} & \textbf{53.05} & 0.221          & 0.211          \\
Base        &                     & 54.25          & 52.45 & \textbf{0.226} & \textbf{0.216} \\
\bottomrule
\end{tabular}

\end{table}

\section{Conclusion and Future work}
In this paper, we have proposed a strategy to perform partial binarization of neural networks in a controlled sense, thereby constructing \textit{budgeted binary neural network (B2NN)}. We presented \texttt{MixBin}, an efficient strategy for finding a well optimized mixture of binary and full precision components in a given network architecture. \texttt{MixBin} allows to explicitly choose the approximate fraction of the network to be kept as binary, thereby presenting the flexibility to adapt the inference cost to a prescribed budget. Numerical experiments conducted on various datasets and model choices support our claim that the B2NNs obtained from our \texttt{MixBin} strategy are significantly better than those obtained from random selection, iterative selection, and even more elegant strategies based on popular methods from the field of model compression. This is strongly evident from the results presented on ImageNet-1K dataset as well as the downstream task of object tracking where significant improvements over the baselines are reported. Overall, from the results and discussions presented earlier in the paper, it can be concluded that \texttt{MixBin} is an efficient and effective strategy for constructing B2NNs that can maximally utilize the available computational resources.


\textbf{Limitations and future work. }Although the presented \texttt{MixBin} strategy results in mixing of full-precision and binary components which are superior over the chosen competitive baselines at similar budgets, there is still scope of improving it. In this paper, we have focused on optimizing the selection process at the layer level. However, there is potential for further investigation into how the network's behavior will be affected by a more granular approach, specifically at the channel level. Additionally, it would be intriguing to explore the application of our layer selection method for transformer-based architectures and assess its efficiency in that context. Moreover, the currently available hardware are limited in terms of fully exploiting the power of binary neural networks, and similar limitation exists for the B2NNs designed using our methodology. However, we believe that with the rapid developments happening in this field, this should be resolved soon.

{\small
\bibliographystyle{unsrt}
\bibliography{egbib}
}

\end{document}



\appendix

\section{Experimental Setup}
\textbf{Configuration of hyperparameters. }cResNet-20, ResNet-18 and MobileNet were trained with batch sizes of 128 at an initial learning rate of $0.1$. For TinyImageNet batch sizes were reduced to 64 and initial learning rate increased to 0.2. We used stochastic gradient descent optimizer with a momentum value of $0.9$ and weight decay of $10^{-4}$, whereas for MobileNet weight decay of $10^{-5}$ was used. Cross Entropy Loss was used with label smoothing of 0.01. We use step learning rate strategy to decay learning rate by $0.1$ after 50\% and 75\% of the total epochs.  For CIFAR100, models were trained for 160 epochs with RandomCrop and RandomHorizontalFlip augmentations whereas for TinyImageNet the number of epochs were reduced to 120, and RandomAffine and RandomHorizontalFlip were used as augmentations. Moreover, for ImageNet, models were trained for 60 epochs with  RandomResizedCrop and RandomHorizontalFlip as augmentations.

Backbone of \textsc{SiamFC} was replaced by the first 3 blocks of ResNet-18.
Model was trained with a batch size of 8 at an initial learning rate of $10^{-3}$ using SGD optimizer with momentum $0.9$ and weight decay $5 \times 10^{-4}$. We use Expontial learning rate schedular with gamma value of 0.59 and final learning rate of $10^{-5}$. All experiments were trained for 50 epochs with early stopping enabled. Please refer to open-source implementation of \url{https://github.com/huanglianghua/siamfc-pytorch} for further details. 

\textbf{Hardware and training time. }Single Nvidia V100 32GB card with 512 GB RAM and a 64-core processor was used for running all the experiments. The training time for pre-training ResNet-18 on ImageNet was 30 hours, and each subsequent run on the specified budget also took 30 hours.


\section{Iterative Selection}

In iterative search strategy we identify the right layers of any given network to be converted to binary or full precision, one-by-one. This approach is described in Algorithm \ref{iter_algo}.
For the $k$ out of $n$ layers to be binarized, the $j^{\text{th}}$ step of binarization, where $j \in [1, k]$, can be stated as finding the optimal layer $\boldsymbol\theta^* \in \boldsymbol\Theta_{(j)}$ to be binarized. It can be mathematically stated as follows.
\begin{algorithm}[H]
\SetAlgoLined
\SetKwInOut{Input}{Input}
\SetKwInOut{Output}{Output}
\SetKwInOut{Given}{Given}
\Given{Empty set \{\};Current layer chosen $\boldsymbol\theta_j$; 
       Optimal layer chosen $\boldsymbol\theta_j^*$;\\
       Current objective function $\mathcal{L}$;
       Optimal objective function $\mathcal{L}^*$;\\
       }
\Input{Network weight set $\boldsymbol\Theta$}
\Output{Binarized weight set $\boldsymbol\Phi$}

$\boldsymbol\Phi \gets \{\}$\\
\For {$j = 1 \hdots k$} 
{
    $\mathcal{L}^* \gets \infty$ \\
    \For{$\boldsymbol\theta_j \in \boldsymbol\Theta$}
    {
        $\mathcal{L} \gets \Call{Iterative Selection}{\boldsymbol\theta_j+\boldsymbol\Phi, \boldsymbol\Theta-\boldsymbol\theta_j,  }$ \\
        \If{$\mathcal{L} < \mathcal{L}^*$}
        {
            $\mathcal{L}^* \gets \mathcal{L}$ \\
            $\boldsymbol\theta_j^* \gets \boldsymbol\theta_j$
        }
    }
    $\boldsymbol\Phi \gets \Call{push}{\boldsymbol\theta_j^*}$\\
    $\boldsymbol\Theta \rightarrow \Call{pop}{\boldsymbol\theta_j^*}$\\
}
\caption{Iterative Selection}
\label{iter_algo}
\end{algorithm}

\begin{equation}
\begin{aligned}
    \boldsymbol\theta^*_{(j)} =  \underset{\boldsymbol\theta \subset \boldsymbol\Theta_{(j)}, \mathbf{W}}{\text{argmin}} \enskip \mathcal{L}(\mathcal{F}(\boldsymbol\theta+ \boldsymbol\Phi_{(j)}, \boldsymbol\Theta_{(j)}-\boldsymbol\theta, \mathbf{x}), \mathbf{y}) \nonumber \\ \text{s.t.} \quad \mathcal{B}(\boldsymbol\theta+ \boldsymbol\Phi_{(j)}, \boldsymbol\Theta_{(j)}-\boldsymbol\theta)\leq \mathcal{B}_0, 
\end{aligned}
\end{equation}

where $\boldsymbol\Theta_{(j)} = \boldsymbol\Theta - \boldsymbol\Phi_{(j)}$. Here, $\boldsymbol\Phi_{(j)}$ denotes the layers that have already been binarized in the previous $j-1$ steps and is defined as $\boldsymbol\Phi_{(j)} = \{\boldsymbol\theta^*_{(1)}, \boldsymbol\theta^*_{(2)}, \hdots, \boldsymbol\theta^*_{(j-1)}\}$, where $\boldsymbol\theta^*_{j}$ denotes the optimal layer chosen at the $j^{\text{th}}$ step of binarization to obtain B2NN. For calculating $\boldsymbol\theta^*_{j}$, we perform brute search over all elements of $\boldsymbol\Theta_{(j)}$ and choose the layer, which when binarized, maximizes the performance of the intermediate B2NN model.

\section{Broader Impact}
We proposed a new paradigm to perform partial binarization wherein layers are either full precision or full binary layers. As highlighted in our experiments, it results in improved model efficiency, enhanced model compression, performance gains in image classification and object tracking, and transferability across datasets. These broader impacts contribute to advancements in efficient and accurate neural networks, enabling their deployment in various resource-constrained scenarios and application domains. Moreover, the reduced footprint of such AI systems is particularly important because the demand for AI continues to grow, and every consumption is becoming a critical concern. Further, the improved inference speed of the devices due to model compression can enhance the user experience and enable new applications in areas such as real-time object tracking, medical diaganosis, among others. Since the B2NN models are light, these also allow sensitive data to be processed locally rather than on cloud servers. This can enhance privacy by minimizing  the sharing of personal data on external servers. 

This highlights the need of more research focused on layer selection and the potential need of specialized hardware which can benefit from B2NNs. Also, binarization, which is one of the strongest compression techniques, will help reduce the hardware and computation constraints which inhibit the use of larger and complex models for production.  Further, as machine learning continues to grow in scale and complexity, the environmental impact of training and inference becomes increasingly significant. Model compression techniques can reduce the computational resources required for training and inference, leading to lower energy consumption and a reduced carbon footprint.

\section{Experiments: Additional Details}
\subsection{Choice of activation function}
\setlength{\tabcolsep}{1.75pt}

\begin{table}[h]
\caption{Performance of B2NNs with different activation function obtained at various FLOPs on CIFAR100 with cResNet20. FLOPs for full precision network is $4.14 \times 10^7$.}
\centering
\resizebox{\columnwidth}{!}{

\begin{tabular}{cccccc}
\toprule
Remaining & \multicolumn{5}{c}{Activation function} \\
\cmidrule(lr){2-6}
\label{tab-activations-out}
 FLOPs (\%) & Identity     & ReLU         & HardTanh     & BinReLU     &RPReLU   \\
\midrule
100.00            & 16.40 & \textbf{65.44} & 61.15 & 64.98 & 63.83 \\
88.78             & 42.72 & 54.66 & 60.64 & \textbf{64.53} & 63.89 \\
77.57             & 48.12 & 54.33 & 60.23 & \textbf{64.72} & 64.4 \\
66.35             & 50.60 & 41.23 & 59.65 & \textbf{64.66} & 63.72 \\
52.33             & 53.13 & 41.28 & 58.87 & \textbf{63.90} & 63.43 \\
32.71             & 53.68 & 29.18 & 54.00 & 61.11 & \textbf{61.65} \\
4.67              & 51.69 & 20.39 & 48.05 & \textbf{53.72} & 53.96 \\
\bottomrule
\end{tabular}}
\end{table}

\textbf{BinReLU activation function. }The proposed BinReLU activation function is designed to enhance the stability of the full-precision as well as binary components of a B2NN in general. Mathematically, BinReLU can be stated as 
\begin{equation}
    f(x) = 
    \begin{cases}
    -1 \text{ if } x \leq -1 \\
    x \text{ otherwise}
    \end{cases},
\end{equation}
where $x$ and $f(x)$ denote the input and output of the activation function. 

Table \ref{tab-activations-out} shows the BinReLU function together with ReLU, HardTanh, RPRelU and Identity mapping. For full-precision networks, ReLU is considered a very effective choice of activation, however, since it eliminates the activation information below 0, it does not work well for binary networks. For BNNs, either of HardTanh or Identity functions are preferred. However, both these activations do not work well for the real-valued networks (Table \ref{tab-activations-out}). Note that an identity mapping works well for BNN since for such cases, nonlinearity is inherently introduced through the squashing of the activation values to -1 and 1 using Sign$(\cdot)$. BinReLU is inspired from the other activations stated here in a sense that it preserves the characteristics of ReLU for positive activations and keeps them real-valued, and also ensures that the activation information between -1 and 0 is preserved.

\subsection{Performance on simple classification datasets. }
This section presents results obtained with \texttt{MixBin}, along with various model compression baselines, on the CIFAR-100 and TinyImageNet datasets. Baselines here include Network Slimming, HAWQ-V2, Adaptive MBQ, BNAS and random selection.

Table \ref{tab-c100-cResNet20} shows performance scores for cResNet-20 for CIFAR-100. Additionally, we provide a sensitivity analysis for this setting, in which we calculate the error in accuracy by conducting three separate runs on different random seeds. Table \ref{tab-mnetv1} and \ref{tab-resnet18} present result on TinyImageNet dataset.

\setlength{\tabcolsep}{2pt}
\begin{table}
\caption{Performance scores for  the MobileNetV1 architecture on TinyImageNet datasets for five compression methods: Network Slimming, HAWQ-V2, Adaptive MBQ, BNAS and \texttt{MixBin} (ours). The term "budget" refers to the percentage of FLOPs that remain after applying the respective compression method.}
\label{tab-mnetv1}
\centering
\resizebox{\columnwidth}{!}{

\begin{tabular}{lcccc}
\toprule
Method & Budget (\%)  &  Acc. (\%) $\uparrow$  & FLOPs (\%) $\downarrow$ &  \\ 
\midrule
Full Precision Network    & - & 52.85 & 100 \\
Binary Network            & - & 34.28 & 4.67 \\
\midrule
Network Slimming          &\multirow{7}{*}{80} & 53.67 & 80\\
HAWQ-V2                   & & 51.86 & 81.03\\
Adaptive MBQ              & & 50.43 & 81.72\\
BNAS                      & & 51.09 & 81.20\\
$\texttt{MixBin}_{Grad}$  & & 56.62 & 80.15\\
$\texttt{MixBin}_{Loss}$  & & 54.26 & 77.03\\
Random                    & & 50.05 & 81.32\\
\midrule
Network Slimming          &\multirow{7}{*}{60} & 51.44 & 60\\
HAWQ-V2                   & & 49.82 & 61.88\\
Adaptive MBQ              & & 49.69 & 61.88\\
BNAS                      & & 49.48 & 61.88\\
$\texttt{MixBin}_{Grad}$  & & 52.11 & 63.18\\
$\texttt{MixBin}_{Loss}$  & & 50.75 & 61.61\\
Random                    & & 49.27 & 61.88\\
\midrule
Network Slimming          &\multirow{7}{*}{40} & 45.83 & 40\\
HAWQ-V2                   & & 46.10 & 41.24\\
Adaptive MBQ              & & 43.03 & 41.03\\
BNAS                      & & 46.48 & 40.08\\
$\texttt{MixBin}_{Grad}$  & & 48.50 & 39.68\\
$\texttt{MixBin}_{Loss}$  & & 47.69 & 38.38\\
Random                    & & 42.3  & 39.45\\
\midrule
Network Slimming          &\multirow{7}{*}{20} & 35.33 & 20\\
HAWQ-V2                   & & 39.31 & 21.43\\
Adaptive MBQ              & & 39.73 &  9.13\\
BNAS                      & & 42.06 & 16.71\\
$\texttt{MixBin}_{Grad}$  & & 41.12 & 13.83\\
$\texttt{MixBin}_{Loss}$  & & 41.12 & 13.83\\
Random                    & & 38.34 &  20.31\\
\bottomrule
\end{tabular}}
\end{table}

\setlength{\tabcolsep}{1.75pt}
\begin{table}[h!]
\caption{Performance scores for  the cResNet-20 architecture on CIFAR-100 datasets for five compression methods: Network Slimming, HAWQ-V2, Adaptive MBQ, BNAS and \texttt{MixBin} (ours). The Full precision network has a total of $4.14\times10^7$ FLOPs, while the binary network generated using the Bi-RealNet method has $1.93\times10^6$ FLOPs. The term "budget" refers to the percentage of FLOPs that remain after applying the respective compression method.}
\label{tab-c100-cResNet20}
\centering
\resizebox{\columnwidth}{!}{

\begin{tabular}{lccccc}
\toprule
Method & Budget(\%)  &  Acc.(\%)$\uparrow$  & FLOPs(\%)$\downarrow$ &  Error$\downarrow$ \\ 
\midrule
Full Precision Network    & - & 64.76 & 100 & 0.48 \\
Binary Network            & - & 53.18 & 4.67 & 0.45\\
\midrule
Network Slimming          &\multirow{7}{*}{80} & 66.67 & 80 & 0.48\\
HAWQ-V2                   & & 65.90 & 77.57 & 0.18\\
Adaptive MBQ              & & 65.77 & 77.57 & 0.15\\
BNAS                      & & 65.57 & 77.57 & 0.22\\
$\texttt{MixBin}_{Grad}$  & & 65.72 & 77.57 & 0.24\\
$\texttt{MixBin}_{Loss}$  & & 66.01 & 77.57 & 0.23\\
Random                    & & 65.03 & 80.37 & 0.53\\
\midrule
Network Slimming          &\multirow{7}{*}{60} & 65.75 & 60 & 0.31\\
HAWQ-V2                   & & 65.77 & 60.75 & 0.38\\
Adaptive MBQ              & & 65.70 & 60.75 & 0.18\\
BNAS                      & & 65.61 & 60.75 & 0.31\\
$\texttt{MixBin}_{Grad}$  & & 65.85 & 60.75 & 0.19\\
$\texttt{MixBin}_{Loss}$  & & 65.62 & 60.75 & 0.39\\
Random                    & & 65.31 & 60.75 & 0.34\\
\midrule
Network Slimming          &\multirow{7}{*}{40} & 63.87 & 40 & 0.14\\
HAWQ-V2                   & & 64.97 & 38.31 & 0.35\\
Adaptive MBQ              & & 61.96 & 38.31 & 2.25\\
BNAS                      & & 61.15 & 38.31 & 1.31\\
$\texttt{MixBin}_{Grad}$  & & 65.32 & 38.31 & 0.42\\
$\texttt{MixBin}_{Loss}$  & & 65.36 & 38.31 & 0.32\\
Random                    & & 61.11 & 43.92 & 0.89\\

\midrule
Network Slimming          &\multirow{7}{*}{20} & 59.20 & 20 & 0.42\\
HAWQ-V2                   & & 62.03 & 21.49 & 0.17\\
Adaptive MBQ              & & 58.46 & 18.69 & 1.93\\
BNAS                      & & 57.59 & 21.49 & 0.19\\
$\texttt{MixBin}_{Grad}$  & & 63.76 & 21.49 & 0.47\\
$\texttt{MixBin}_{Loss}$  & & 63.49 & 21.49 & 0.23\\
Random                    & & 58.56 & 21.49 & 0.63\\
\midrule
Network Slimming          &\multirow{7}{*}{10} & 52.52 & 10 & 0.39\\
HAWQ-V2                   & & 58.41 & 13.12 & 0.31\\
Adaptive MBQ              & & 56.86 & 10.28 & 1.81\\
BNAS                      & & 57.03 & 10.28 & 0.50\\
$\texttt{MixBin}_{Grad}$  & & 60.27 & 11.08 & 0.50\\
$\texttt{MixBin}_{Loss}$  & & 57.54 & 10.28 & 0.75\\
Random                    & & 56.47 & 10.28 & 0.59\\
\bottomrule
\end{tabular}}
\end{table}

\setlength{\tabcolsep}{1.75pt}
\begin{table}[h!]
\caption{Performance scores for  the ResNet-18 architecture on TinyImageNet datasets for five compression methods: Network Slimming, HAWQ-V2, Adaptive MBQ, BNAS and \texttt{MixBin} (ours). The Full precision network has a total of $5.63\times10^8$ FLOPs. The term "budget" refers to the percentage of FLOPs that remain after applying the respective compression method.}
\label{tab-resnet18}
\centering
\resizebox{\columnwidth}{!}{

\begin{tabular}{lcccc}
\toprule
Method & Budget (\%)  &  Acc. (\%) $\uparrow$  & FLOPs (\%) $\downarrow$ &  \\ 
\midrule
Full Precision Network    & - & 56.88 & 100 \\
Binary Network            & - & 44.43 & 4.67 \\
\midrule
Network Slimming          &\multirow{7}{*}{80} & 55.45 & 80\\
HAWQ-V2                   & & 57.58 & 81.25\\
Adaptive MBQ              & & 57.55 & 81.25\\
BNAS                      & & 56.99 & 81.25\\
$\texttt{MixBin}_{Grad}$  & & 57.11 & 81.25\\
$\texttt{MixBin}_{Loss}$  & & 57.45 & 81.25\\
Random                    & & 56.89 & 81.25   \\
\midrule
Network Slimming          &\multirow{7}{*}{60} & 55.3 & 60\\
HAWQ-V2                   & & 56.80 & 62.50\\
Adaptive MBQ              & & 56.71 & 62.50\\
BNAS                      & & 57.06 & 62.50\\
$\texttt{MixBin}_{Grad}$  & & 56.67 & 62.50\\
$\texttt{MixBin}_{Loss}$  & & 57.51 & 62.50\\
Random                    & & 56.6 &  62.50  \\
\midrule
Network Slimming          &\multirow{7}{*}{40} & 54.12 & 40\\
HAWQ-V2                   & & 55.70 & 43.76\\
Adaptive MBQ              & & 56.10 & 43.76\\
BNAS                      & & 55.71 & 43.76\\
$\texttt{MixBin}_{Grad}$  & & 56.44 & 43.76\\
$\texttt{MixBin}_{Loss}$  & & 56.18 & 43.76\\
Random                    & & 56.0 &  40.62   \\

\midrule
Network Slimming          &\multirow{7}{*}{20} & 54.00 & 20\\
HAWQ-V2                   & & 54.36 & 21.88\\
Adaptive MBQ              & & 54.58 & 25.01\\
BNAS                      & & 53.69 & 21.88\\
$\texttt{MixBin}_{Grad}$  & & 55.08 & 21.88\\
$\texttt{MixBin}_{Loss}$  & & 54.86 & 21.88\\
Random                    & & 54.25 & 24.12    \\
\midrule
Network Slimming          &\multirow{7}{*}{10} & 51.83 & 10\\
HAWQ-V2                   & & 53.57 & 12.51\\
Adaptive MBQ              & & 53.87 & 12.51\\
BNAS                      & & 53.43 & 12.51\\
$\texttt{MixBin}_{Grad}$  & & 54.25 & 12.51\\
$\texttt{MixBin}_{Loss}$  & & 54.25 & 12.51\\
Random                    & & 54.02 & 10.92    \\
\bottomrule
\end{tabular}}
\end{table}

\subsection{Effect of binarizing different parts of a network.}
The performance of the constructed B2NN model depends heavily on the choice of $\boldsymbol\Phi$. In Table \ref{tab-diff-parts}, we compare the performance of MixBin with several trivial baselines.

\setlength{\tabcolsep}{1.75 pt}

\begin{table}[h]
\caption{Performance comparision of B2NNs obtained using \texttt{MixBin} vs. the various trivial approaches. The B2NNs are constructed using cResNet-20 architecture on CIFAR-100 datasets. Here, \emph{Rear-BN} and \emph{Front-BN} represent B2NNs constructed with binary layer placed in the rear and front parts of the network, respectively. \emph{Pseudo-random} involves modifying 30\% of the MixBin generated B2NNs. The Full Precision network and Binary Network have $4.14\times10^7$ and $1.93\times10^6$ FLOPs respectively.}
\label{tab-diff-parts}
\centering
\resizebox{\columnwidth}{!}{

\begin{tabular}{lccccc}

\toprule
Method & Budget(\%)  &  Acc.(\%)$\uparrow$  & FLOPs(\%)$\downarrow$ &  Error$\downarrow$\\ 
\midrule
Full Precision Network    & - & 64.76 & 100 & 0.48\\
Binary Network            & - & 53.18 & 4.67 & 0.45\\
\midrule
\emph{Front-BN}          &\multirow{5}{*}{80} & 65.12 & 83.17 & 0.52\\
\emph{Rear-BN}            & & 63.51 & 83.17 & 1.31\\
\emph{Pseudo-random}      & & 65.18 & 83.17 & 0.07\\
$\texttt{MixBin}_{Grad}$  & & 65.72 & 77.57 & 0.24\\
$\texttt{MixBin}_{Loss}$  & & 66.01 & 77.57 & 0.23\\
\midrule
\emph{Front-BN}          &\multirow{5}{*}{60} & 64.13 & 63.54 & 0.37\\
\emph{Rear-BN}            & & 59.32 & 63.54 & 0.02\\
\emph{Pseudo-random}      & & 64.60 & 63.54 & 0.22\\
$\texttt{MixBin}_{Grad}$  & & 65.85 & 60.75 & 0.19\\
$\texttt{MixBin}_{Loss}$  & & 65.62 & 60.75 & 0.39\\
\midrule
\emph{Front-BN}          &\multirow{5}{*}{40} & 63.64 & 41.12 & 0.59\\
\emph{Rear-BN}            & & 57.78 & 41.12 & 0.43\\
\emph{Pseudo-random}      & & 64.28 & 38.31 & 0.62\\
$\texttt{MixBin}_{Grad}$  & & 65.32 & 38.31 & 0.42\\
$\texttt{MixBin}_{Loss}$  & & 65.36 & 38.31 & 0.32\\

\midrule
\emph{Front-BN}          &\multirow{5}{*}{20} & 61.79 & 21.49 & 0.66\\
\emph{Rear-BN}            & & 56.49 & 21.49 & 0.60\\
\emph{Pseudo-random}      & & 61.40 & 18.68 & 0.49\\
$\texttt{MixBin}_{Grad}$  & & 63.76 & 21.49 & 0.47\\
$\texttt{MixBin}_{Loss}$  & & 63.49 & 21.49 & 0.23\\
\midrule
\emph{Front-BN}          &\multirow{5}{*}{10} & 59.28 & 10.28 & 0.41\\
\emph{Rear-BN}            & & 55.91 & 10.28 & 0.56\\
\emph{Pseudo-random}      & & 55.28 & 10.28 & 0.57\\
$\texttt{MixBin}_{Grad}$  & & 60.27 & 11.08 & 0.50\\
$\texttt{MixBin}_{Loss}$  & & 57.54 & 10.28 & 0.75\\
\bottomrule
\end{tabular}}
\vspace{-4mm}

\end{table}